\documentclass[10pt,twocolumn,letterpaper]{article}

\usepackage{3dv}
\usepackage{times}
\usepackage{epsfig}
\usepackage{subfigure}
\usepackage{graphicx}
\usepackage{amsmath}
\usepackage{amssymb}
\usepackage{bm}

\makeatletter
\newcommand{\xLeftrightarrow}[2][]{\ext@arrow 0099\Leftrightarrowfill@{#1}{#2}}
\makeatother

\usepackage[pagebackref=true,breaklinks=true,letterpaper=true,colorlinks,bookmarks=false]{hyperref}

\threedvfinalcopy 

\def\threedvPaperID{155} 

\ifthreedvfinal\fi
\begin{document}

\title{Online Stability Improvement of\\Gr\"obner Basis Solvers using Deep Learning}

\author{$^{\text{1,2)}}$Wanting Xu \hspace{0.5cm} $^{\text{1,2)}}$Lan Hu \hspace{0.5cm} $^{\text{2)}}$Manolis C. Tsakiris \hspace{0.5cm} $^{\text{1,2)}}$Laurent Kneip\\
$^{\text{1)}}$Mobile Perception Lab, $^{\text{2)}}$ShanghaiTech\\
{\tt\small \{xuwt,hulan,mtsakiris,lkneip\}@shanghaitech.edu.cn}}

\maketitle
\thispagestyle{empty}

\begin{abstract}
Over the past decade, the Gr\"obner basis theory and automatic solver generation have lead to a large number of solutions to geometric vision problems. In practically all cases, the derived solvers apply a fixed elimination template to calculate the Gr\"obner basis and thereby identify the zero-dimensional variety of the original polynomial constraints. However, it is clear that different variable or monomial orderings lead to different elimination templates, and we show that they may present a large variability in accuracy for a certain instance of a problem. The present paper has two contributions. We first show that for a common class of problems in geometric vision, variable reordering simply translates into a permutation of the columns of the initial coefficient matrix, and that---as a result---one and the same elimination template can be reused in different ways, each one leading to potentially different accuracy. We then prove that the original set of coefficients may contain sufficient information to train a classifier for online selection of a good solver, most notably at the cost of only a small computational overhead. We demonstrate wide applicability at the hand of generic dense polynomial problem solvers, as well as a concrete solver from geometric vision.
\end{abstract}

\section{Introduction}

Geometric closed-form solvers represent the cornerstones of structure from motion, as they permit the initialization of both intrinsic \cite{kukelova13} or extrinsic camera parameters \cite{kneip2014upnp,stewenius2006recent} when no prior information is available. Given algebraic incidence relationships, the derivation of a closed-form solver usually starts by applying variable elimination techniques, thus leading to an initial system of polynomial constraints in typically few unknowns. A meanwhile established methodology of solving such polynomial systems of equations relies on algebraic geometry and the Gr\"obner basis method. The present paper addresses the stability and accuracy of Gr\"obner basis solvers.

In simple terms, the Gr\"obner basis method proceeds by iteratively generating new polynomials (so-called \textit{Syzygies}) vanishing on the original variety, each time adding them to the set of ideal generators if their reduction by the already existing generators does not lead to a zero remainder. The procedure is repeated until all possible polynomial reductions have taken place. The success of the Gr\"obner basis method relies on the insight that the expensive search for the basis does not need to be repeated for each new instance of a certain type of problem; the sequence of the generated polynomials remains the same for each \textit{general} instance of a problem. As a result, efficient solvers are generated by translating the sequence of polynomial generations into an \textit{elimination template}. At online stage, the elimination template is filled with all initial polynomials of a specific problem instance as well as their required multiplications by monomials, and then subjected to for example a Gauss-Jordan elimination. For more details on the Gr\"obner basis method as well as automatic solver generation, the reader is kindly referred to \cite{cox1994ideals}, \cite{cox2006using} and \cite{kukelova2008automatic}. Here we simply note that the elimination template for a particular problem typically remains fixed at online stage.

It is well-known that the complexity of finding a Gr\"obner basis depends on a number of factors. To start with, an effort has to be made to find a good parametrization for which the order of the equations and the remaining number of unknowns are kept as low as possible. The next step consists of finding suitable variable and monomial orderings for efficient retrieval of the Gr\"obner basis (and thus a compact size of the elimination template). Even though there are infinitely many monomial orderings, there are only finitely many (reduced) Gr\"obner bases corresponding to a given set of equations \cite{mora88}. Among these, the one corresponding to the graded-reverse lexicographical (\textit{grevlex}) ordering is typically considered a good choice. On the other hand, \cite{larsson2018beyond} considers searching among all these finitely many Gr\"obner bases in pursuit of an optimal choice. The primary motivation of this work is given by the fact that---for one and the same instance of a certain type of problem---there may be different elimination templates with comparable computational complexity but substantial differences in the numerical stability of the retrieved solutions. Deciding for a single elimination template at offline stage therefore appears as a weakness, and a smart way of picking one of several comparable elimination templates at online stage may lead to substantial benefits in terms of the robustness and accuracy of the solutions. We have two main contributions:
\begin{itemize}
\item We demonstrate that simple variable reordering potentially has a significant impact on the robustness and accuracy of the solutions. We furthermore show that for a large class of common polynomial problems (including dense polynomial systems) variable reordering simply translates into a permutation of the columns of the original coefficient matrix, thus enabling one and the same elimination template to be used in as many ways as there are permutations of the input variables. We furthermore demonstrate that such column permutations potentially have a significant impact on the quality of the solutions. The result applies to a large class of polynomial problems that includes many 3D rotation-based solvers in geometric vision.
\item We demonstrate that the coefficients of the original set of polynomials do in fact contain sufficient information to train an artificial neural network that is able to predict which of many possible elimination templates will lead to a stable result. The networks are typically small and their inference represents an insignificant computational overhead compared to the actual elimination template. In other words, they are amenable to an online selection of the elimination template. In the case of permutation invariant elimination templates, they predict which column permutation to choose.
\end{itemize}
We demonstrate the viability of the approach and its potential to improve solver stability at the hand of general dense polynomial problems as well as a popular solver for finding the absolute pose of a camera from a single image. The paper is organized as follows. Section \ref{sec:theory} introduces back-ground theory and the permutation invariance of elimination templates for a certain class of problems. Section \ref{sec:learning} then depicts the details of our online classifier including training from purely synthetic data. Section \ref{sec:results} finally presents our results on multiple solvers, followed by a brief discussion.

\subsection{Related work}

The Gr\"obner basis theory largely relies on the original work of Bruno Buchberger \cite{buchberger85}, who introduced the well-known \textit{Buchberger algorithm} for the computation of a Gr\"obner basis. Good descriptions of the material can be found in Cox et al.'s introductions to algebraic geometry \cite{cox1994ideals,cox2006using}. One of the pioneering works employing a Gr\"obner basis solver in computer vision is presented by Stewenius et al. \cite{stewenius05}, who apply the technique to derive a closed-form solution to the calibrated generalized relative pose problem. While the application of the Gr\"obner basis method originally involves a manual search for the elimination template, a major breakthrough has been achieved by Kukelova's work on automatic solver generation \cite{kukelova2008automatic}. The method has since been used exhaustively to solve both absolute \cite{zheng13,kneip2014upnp} and relative camera pose estimation problems \cite{stewenius05,kneip12}. The method has furthermore been employed to solve a large variety of more specialized solvers that for example consider the partially uncalibrated or planar case \cite{kukelova13}, directional correspondences \cite{lee14}, or even special geometric arrangements such as two intersecting lines \cite{zhao18}. Gr\"obner basis solvers are important as they utilize a minimal set of points, and thus benefit robust hypothesis and test schemes.

Recent years have shown a number of works aiming at an improvement of solver efficiency with respect to the original solvers generated by Kukelova et al.'s toolbox \cite{kukelova2008automatic}. For example, Bujnak et al.'s main contribution in \cite{bujnak2012making} consists of using a modification of \textit{FGLM} \cite{faugere93} to transform a \textit{grevlex} Gr\"obner basis into a lexicographical one. Solutions to the latter can notably be recovered by efficiently finding the roots of a univariate polynomial. Kukelova et al. \cite{kukelova14} later present an improvement of the reduction of the actual elimination template by exploiting the fact that it often has a sparse, block-diagonal structure. Further improvements are possible in situations in which there is a p-fold symmetry in the variety \cite{ask12} or in which the ideals are saturated \cite{larsson17iccv2}. The most recent advancements that directly address the automatic solver generation issue are presented by Larsson et al. They first present an improved automatic solver generator \cite{larsson2017efficient}, and then explore Gr\"obner fans for a variety of basis choices or even a random sampling scheme of linearly independent monomials in the quotient ring \cite{larsson2018beyond}. Depending on the chosen basis monomials, the solver will notably employ different ideal generators from the elimination template, which potentially leads to improved accuracy or even computational efficiency. The latter contribution is related to our work in that it acknowledges the existence of multiple possible elimination templates. However, similar to most prior art, the choice of the basis and the resulting elimination template is fixed offline at solver generation stage, which limits the flexibility of the approach. One notable exception that is highly related to our work is presented by Byr\"od et al. \cite{byrod09}. It introduces online strategies for an improved construction of the action matrix.

To the best of our knowledge, the present work is the first to exploit variable permutations and permutation-invariant polynomial forms to change the behavior of an elimination template at online stage. Furthermore, while neural networks have been recently used to learn permutations (e.g. visual permutation learning \cite{cruz18}), to the best of our knowledge, we are the first to combine deep learning and algebraic geometry, and devise an automatically trained classifier for efficient online selection of a suitable permutation in the context of polynomial solving.

\section{Permutation invariant polynomial systems}
\label{sec:theory}

The present paper looks at a special class of polynomial systems for which the \textit{support}\footnote{The \textit{support} of a polynomial is given by the set of its monomials.} is invariant with respect to variable permutations. In the continuation, we describe such systems as \textit{permutation invariant}. The present section introduces some necessary notations and defines permutation invariant polynomials and polynomial systems. After demonstrating the potential impact of different variable orderings on solver stability, we will then see how---in the case of permutation invariant polynomials---variable reordering can be translated into column permutations of the original coefficient matrix, thus enabling one and the same elimination template to be reused in different ways, notably one for each variable ordering.

\subsection{Notations}

Let $\mathbf{x}=\{x_1,\ldots,x_n\}$ be the set of variables involved in a polynomial problem defined over the polynomial ring $\mathbb{C}[\mathbf{x}]$. Let $\{f_1,\ldots,f_m\}$, $f_j \in \mathbb{C}[\mathbf{x}]$ furthermore be the original set of polynomials defining the ideal $I=\{\sum_jh_jf_j ~ \vert ~ h_j \in \mathbb{C}[\mathbf{x}]\}$ for which we want to retrieve the zero-dimensional variety $\{\mathbf{x}\in\mathbb{C}^{n}, \text{ s.t. } f_j(\mathbf{x})=0, j=1,\ldots,m\}$\footnote{Note that we make the assumption that---at least in the complex field $\mathbb{C}$---the ideal properly defines a zero-dimensional variety, i.e. a non-empty finite set of points.}. Each polynomial is of the form
\begin{equation}
    f_j = \sum_{i=1}^{k} c_{ji} \mathbf{x}^{\bm{\alpha}_i}, \label{eq:f}
\end{equation}
where $c_{ji}$ is assumed to be a coefficient drawn from the field of real numbers $\mathbb{R}$, and $\bm{\alpha}=\{\alpha_1,...,\alpha_n\}$ denotes the set of exponents of a particular monomial such that
\begin{equation}
    \mathbf{x}^{\bm{\alpha}}=x_1^{\alpha_1}\cdots x_n^{\alpha_n}.
\end{equation}

In a slight abuse of notation, $\mathbf{x}$ and $\bm{\alpha}$ will in the following be used interchangeably to denote either the ordered set as defined above, or a column vector composed of the same elements in the same order. Now let $\mathbf{P}$ be an $n\times n$ permutation matrix representing a permutation $\pi$ of $n$ symbols. Then for a polynomial $f_j$ as in \eqref{eq:f}, we can obtain a new polynomial, denoted $\pi(f_j)$, by permuting the multi-exponents of $f_j$ according to $\mathbf{P}$, i.e., 
\begin{equation}
    \pi(f_j) = \sum_{i=1}^{k} c_{ji} \mathbf{x}^{\bm{\mathbf{P}\alpha}_i}. \label{eq:pi(f)}
\end{equation} \footnote{We note here that this is equivalent to a permuting the polynomial variables $x_1,\ldots,x_n$ in $f_j$ according to the inverse permutation $\pi^{-1}$.}

\subsection{Permutation-invariant polynomials}

\noindent\textit{Definitions}: Let $\mathbf{m}=\{\mathbf{x}^{\bm{\alpha}_1},\ldots,\mathbf{x}^{\bm{\alpha}_k}\}$ be the support of the polynomial. We define a polynomial in variables $\mathbf{x}=\{x_1,\ldots,x_n\}$ to be \textit{permutation-invariant} under a permutation matrix $\mathbf{P}_{n\times n}$ if and only if every element of
\begin{equation}
    \mathbf{m}'=\{\mathbf{x}^{\mathbf{P}\bm{\alpha}_1},\ldots,\mathbf{x}^{\mathbf{P}\bm{\alpha}_k}\}
\end{equation}
is also contained in the original set $\mathbf{m}$, i.e. $\mathbf{x}^{\mathbf{P}\bm{\alpha}_i}\in\{\mathbf{x}^{\bm{\alpha}_1},\ldots,\mathbf{x}^{\bm{\alpha}_k}\},\forall i$. Similarly, we define a polynomial system to be permutation invariant if all composing polynomials are permutation invariant for themselves, and a set of exponent vectors $\mathbf{o}=\{\bm{\alpha}_1,\ldots,\bm{\alpha}_k\}$ is permutation invariant if $\mathbf{P}\bm{\alpha}_i\in\{\bm{\alpha}_1,\ldots,\bm{\alpha}_k\},\forall i$.

The following are a few important examples of polynomials which all comply with this definition:
\begin{itemize}
    \item \textit{Symmetric polynomials}: Polynomials that simply do not change under a permutation. For example, given the polynomial $f(\mathbf{x})=x_1^2 + x_2^2 + x_3^2 + x_4^2 + 1$ and a permutation $\mathbf{P}=\tiny{\left[\begin{matrix}0 & 0 & 1 & 0\\0 & 1 & 0 & 0\\ 0 & 0 & 0 & 1\\1 & 0 & 0 & 0\end{matrix}\right]}$, we obtain $f(\mathbf{x})\xLeftrightarrow{\mathbf{x}^{\bm{\alpha}_i}\leftarrow \mathbf{x}^{\mathbf{P}\bm{\alpha}_i}} x_4^2 + x_2^2 + x_1^2 + x_3^2 + 1$. The resulting polynomial is the same.
    \item \textit{Dense polynomials}: Polynomials in which all monomials up to a certain degree appear. For example, given the polynomial
    $f(\mathbf{x})=c_1x_1^2 + c_2x_1x_2 + c_3x_2^2 + c_4x_1 + c_5x_2 + c_6$ and a permutation $\mathbf{P}=\tiny{\left[\begin{matrix}0 & 1 \\ 1 & 0\end{matrix}\right]}$, we obtain $f(\mathbf{x})\xLeftrightarrow{\mathbf{x}^{\bm{\alpha}_i}\leftarrow \mathbf{x}^{\mathbf{P}\bm{\alpha}_i}} c_1x_2^2 + c_2x_2x_1 + c_3x_1^2 + c_4x_2 + c_5x_1 + c_6$. While the coefficients of identical monomials are changing, the support of the polynomials remains unchanged. An example application that features both dense and symmetric polynomials is given by shuffled linear regression \cite{tsakiris18,peng19}.
    \item \textit{Degree-wise dense polynomials}: Essentially same as dense polynomials, except that they do not necessarily contain all possible monomials up to a certain degree, but simply all possible monomials of certain degrees.
    \item \textit{Special cases}: Consider the bi-variate examples $f_1(\mathbf{x})=c_1x_1^2 + c_2x_2^2 + c_3x_3^2 + 1\in\mathbb{C}[x_1,x_2,x_3]$ and $f_2(\mathbf{x})=c_1x_1^2x_2+c_2x_1x_2^2\in\mathbb{C}[x_1,x_2]$. According to the above definition, they can also be defined as permutation-invariant. The coefficients of the monomials may change while supports remain unchanged.
\end{itemize}

\subsection{Impact of variable reordering}

\begin{figure}[t]
\begin{center}
  \includegraphics[width=0.8\linewidth]{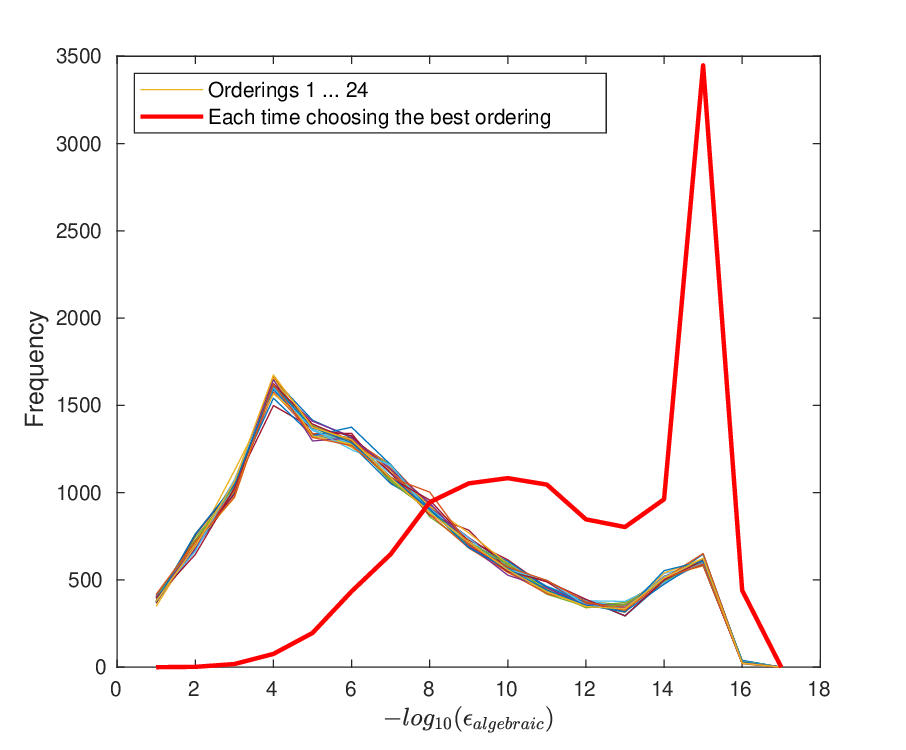}
\end{center}
  \caption{Error distribution over many random instances of the UPnP problem. Each curve represents the distribution obtained by one of the variable orderings. The red curve furthermore indicates the obtainable result if---for each instance---the best variable ordering is chosen.}
\label{fig:example}
\end{figure}

Before we can explain the relevance of permutation-invariance in the context of a Gr\"obner basis solver, we first need to understand the potential impact of variable reordering on the numerical conditioning of the problem. Suppose we are given the following polynomial problem of 3 equations in 3 unknowns, and notably in grevlex ordering based on a variable ordering of $x_1>x_2>x_3$:
\begin{equation}
    \mathbf{c}_1x_3^2 + \mathbf{c}_2x_2^2 + \mathbf{c}_3x_1^2+\mathbf{c}_4x_3+\mathbf{c}_5x_2+\mathbf{c}_6x_1+\mathbf{c}_7=0,
    \label{eq:exampleBeg}
\end{equation}
where $\mathbf{c}_i$ are $3\times 1$ vectors of coefficients. Generating a Gr\"obner basis solver for this problem would lead to a certain elimination template. Choosing the different variable ordering $x_2>x_1>x_3$ will lead to the system
\begin{equation}
    \mathbf{c}_1x_3^2 + \mathbf{c}_3x_1^2 + \mathbf{c}_2x_2^2+\mathbf{c}_4x_3+\mathbf{c}_6x_1+\mathbf{c}_5x_2+\mathbf{c}_7=0.
    \label{eq:example}
\end{equation}
Let us ignore for a moment that the supports remain unchanged due to the permutation-invariance property. Changing the variable ordering generally leads to a different Gr\"obner basis and elimination template, and therefore also numerical behavior. As a concrete example, let us consider the geometric vision problem presented in \cite{kneip2014upnp}. The constraints are given by 4 at most cubic polynomials in 4 unknowns, and the original solver relies on a single elimination template of dimension $141\times 149$. By applying all possible 24 variable orderings and reducing the corresponding elimination templates, we can obtain 24 results for each instance of a problem. Figure \ref{fig:example} finally illustrates error distributions for many random problem instances for each of the 24 possible variable orderings. It furthermore shows one more error distribution of a solution where---for each problem instance---the best of the 24 variable orderings is selected. As expected, on the average each permutation is equally good. However, for each coefficient instance, selecting the best permutation has a clear impact on the quality of the solution.

\subsection{Variable reordering and permutations}

Before proceeding, let us introduce further notations which are closely related to the elimination template itself. The coefficients of our polynomials $f_j, j=1,\ldots,m$ may be grouped in the rows of a coefficient matrix $\mathbf{C}_{m\times h}$, where $h$ represents the number of distinct monomials appearing in the entire system of equations. We may also redefine $\mathbf{m}$ as a column-vector containing these monomials, as in
\begin{equation}
    \mathbf{m} = \left[\mathbf{x}^{\bm{\alpha}_1} \ldots \mathbf{x}^{\bm{\alpha}_h}\right]^{T}.
\end{equation}
Each column $\mathbf{c}_i$ of $\mathbf{C}$ therefore contains the (potentially partially zero) coefficients of monomial $\mathbf{x}^{\bm{\alpha}_i}, i=1,\ldots,h$, which lets us rewrite the original polynomial constraints as
\begin{equation}
    \mathbf{C}\mathbf{m}=\mathbf{0}.
\end{equation}
To conclude, we assume that the ordering of the monomials in $\mathbf{m}$ obeys a total monomial ordering (e.g. grevlex), and also redefine the set $\mathbf{o}=\{\bm{\alpha}_1,\ldots,\bm{\alpha}_h\}$ such that it contains the exponent vectors of all monomials, i.e. the ones in $\mathbf{m}$. Note that the rows and columns of $\mathbf{C}$ are in fact a subset of the rows and columns in the overall elimination template, and they contain all original coefficients.
\begin{figure}[b]
    \centering
    \includegraphics[width=0.90\columnwidth]{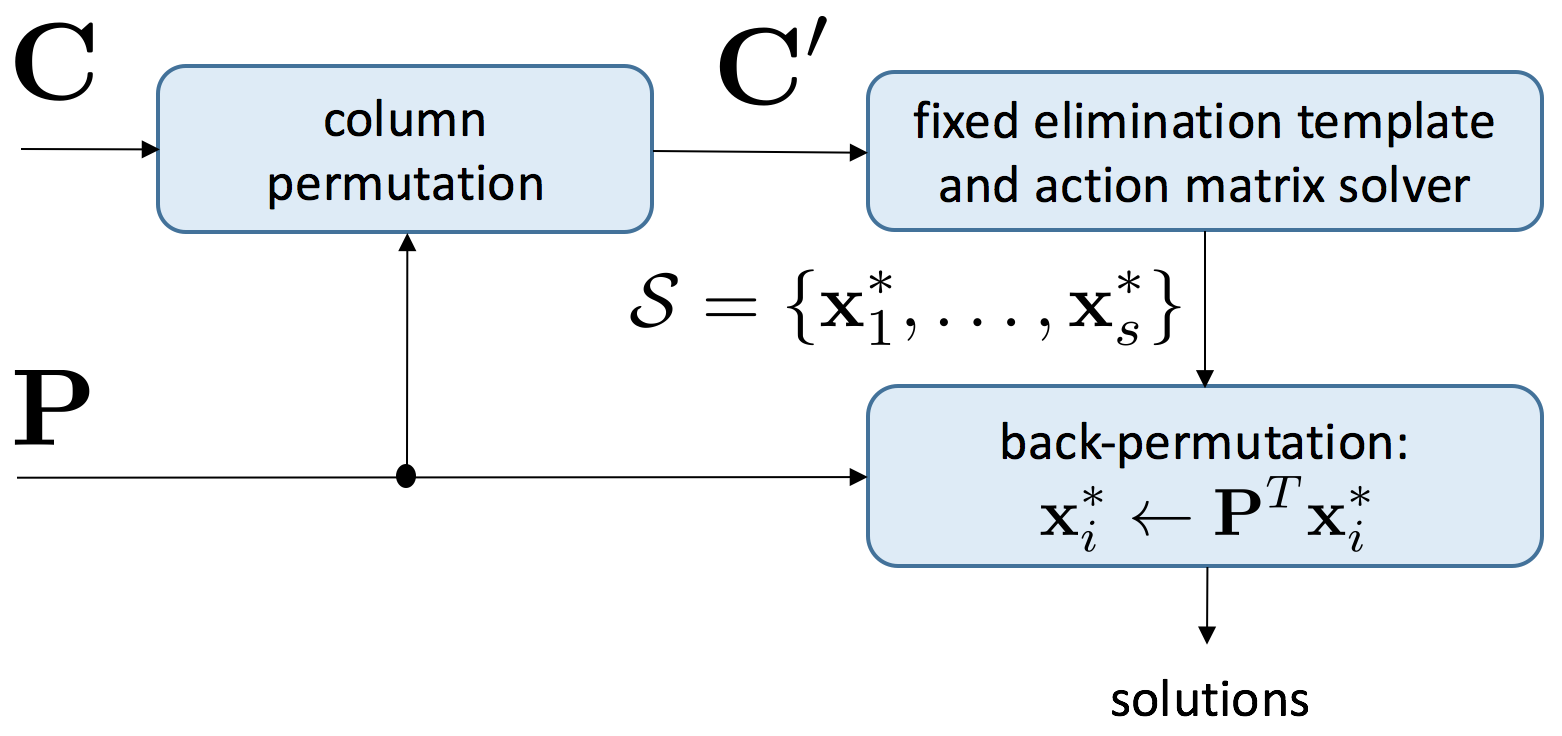}
    \caption{Permutation of the coefficient matrix $\mathbf{C}$ and back-permutation of the solutions $\mathcal{S}=\{\mathbf{x}_1^{*},\ldots,\mathbf{x}_s^{*}\}$
 for a potential improvement of the numerical stability of an elimination template.}
    \label{fig:basicconcept}
\end{figure}

Our main insight is given by the fact that---in the case of permutation invariant polynomials---the elimination templates corresponding to all variable orderings are in fact the same. This is easily recognized by applying a simple permutation of the unknowns to undo the variable reordering. Taking (\ref{eq:example}) as an example, applying the permutation $\mathbf{P}=\tiny{\left[\begin{matrix} 0 & 1 & 0\\ 1 & 0 & 0\\0& 0& 1\end{matrix}\right]}$ to $\mathbf{x}$ would re-establish a polynomial system of exactly same form and order as (\ref{eq:exampleBeg}), except that the coefficient column-vectors would have been permuted. The same remains true more generally for any coefficient matrix $\mathbf{C}$ for which the set $\mathbf{o}$ is permutation invariant. Note that---in accordance with our definition of a permutation invariant polynomial system---if all individual polynomials $f_i$ are permutation invariant, the set $\mathbf{o}$ must be so as well. As illustrated in Figure \ref{fig:basicconcept}, it is therefore possible to solve a permutation invariant polynomial system in as many ways as there are variable permutations. All that needs to be done is a back-permutation of each identified solution. The important point is that, despite the use of only a single elimination template, different permutations potentially lead to different numerical stability pretty much in the same way different matrix factorisations in linear algebra would yield different numerical solutions to the same system of linear equations.

The remaining question is how to permute the columns of the coefficient matrix. Let $\mathbf{C}'=[\mathbf{c}'_1 \ldots \mathbf{c}'_h]$ be the target coefficient matrix with the permuted columns, and $\mathbf{P}$ be the permutation matrix. It can be obtained from the original coefficient matrix $\mathbf{C}$ by applying
\begin{equation}
    \mathbf{c}'_i = \mathbf{c}_{\operatorname{findindex}(\mathbf{P}^T\bm{\alpha}_i,\mathbf{o})},
\end{equation}
where $\operatorname{findindex}(\bm{\alpha},\mathbf{o})$ is a function that returns the index of $\bm{\alpha}$ inside the set $\mathbf{o}$.

\section{Online selection via deep learning}
\label{sec:learning}

The previous section explained how for permutation invariant polynomial systems, a single elimination template can be used in many ways to retrieve the solution, each one potentially leading to different numerical accuracy. The present section addresses the question of how a good permutation for a particular instance of a problem can be found upfront with only very little computational overhead. We start by seeing the basic idea which consists of applying a classifier that is able to select a good permutation at online stage. The remainder of the section then addresses the question of how to train such a classifier.

\subsection{Basic approach}

\begin{figure}[b]
    \centering
    \includegraphics[width=0.9\columnwidth]{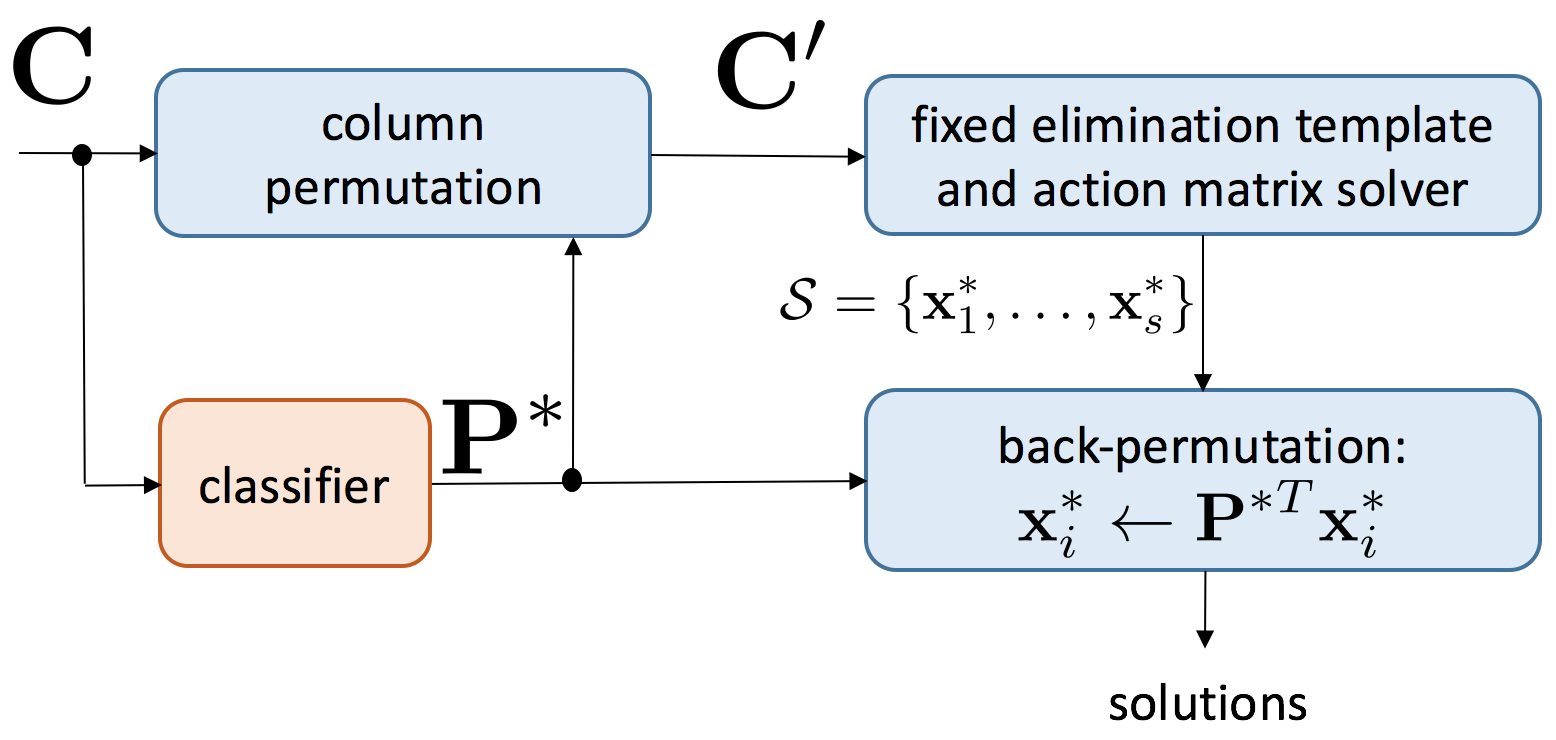}
    \caption{Permutation-aware Gr\"obner basis solver with added classifier able to predict a good permutation $\mathbf{P}^{*}$ directly from the original coefficient matrix $\mathbf{C}$.}
    \label{fig:withclassifier}
\end{figure}

The basic idea consists of simply adding a classifier which is able to predict a good permutation $\mathbf{P}^{*}$ directly from the original coefficients. The modified flow-chart of the solver is depicted in Figure \ref{fig:withclassifier}.

\begin{figure}[t]
    \centering
    \includegraphics[width=\columnwidth]{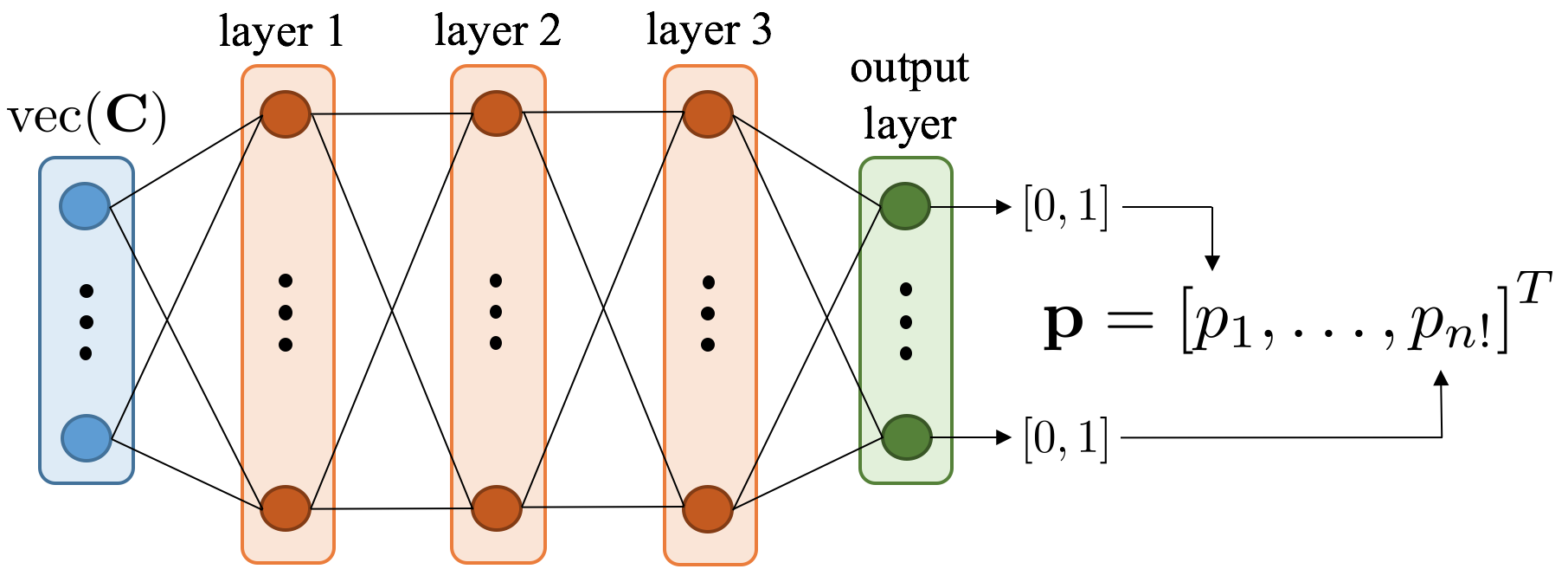}
    \caption{Architecture of our neural network for predicting and selecting a good permutation.}
    \label{fig:network}
\end{figure}

The classifier is a simple four-layer neural network that takes the vectorized coefficients $\operatorname{vec}(\mathbf{C})$ as an input, and produces $n!$ output signals, each one approximating the rank of a certain permutation by a number between 0 and 1 (0=worst, 1=best). The architecture of the network is illustrated in Figure \ref{fig:network}. All layers are fully connected. Each of the three latent layers contains 500 neurons, and therefore has an output dimension of 500. Input dimensions are 500, except for the first layer, which contains $m\times h$ input variables. All activation functions are set to rectified linear units, and batch normalization is added during training. The output layer has 500 inputs and $n!$ outputs, and uses sigmoid activation functions to enforce rank numbers $p_i\in[0,1]$. The outputs are concatenated into a float vector $\mathbf{p}=\left[p_1 \ldots p_{n!}\right]^T$. The chosen permutation $\mathbf{P}^{*}$ is the one corresponding to the largest element in $\mathbf{p}$.

\subsection{Training procedure}

The training procedure is relatively straightforward and relies on synthetic training dataset generation. For a given polynomial problem in $n$ variables, we will start by sampling coefficient matrices $\mathbf{C}$. Note that in practically all scenarios, this process can be done very efficiently. For example, if talking about a polynomial problem with arbitrary independent coefficients, the input coefficient matrix $\mathbf{C}$ can simply be chosen randomly. A more complicated case may be given by camera calibration problems for which the coefficients are no longer fully independent. However, it is easy to generate valid coefficient matrices through automatic simulators that start from random scenes and random camera calibration parameters, and then apply forward projection to find geometrically consistent coefficients.

\begin{figure*}
  \centering
  \includegraphics[width=\textwidth]{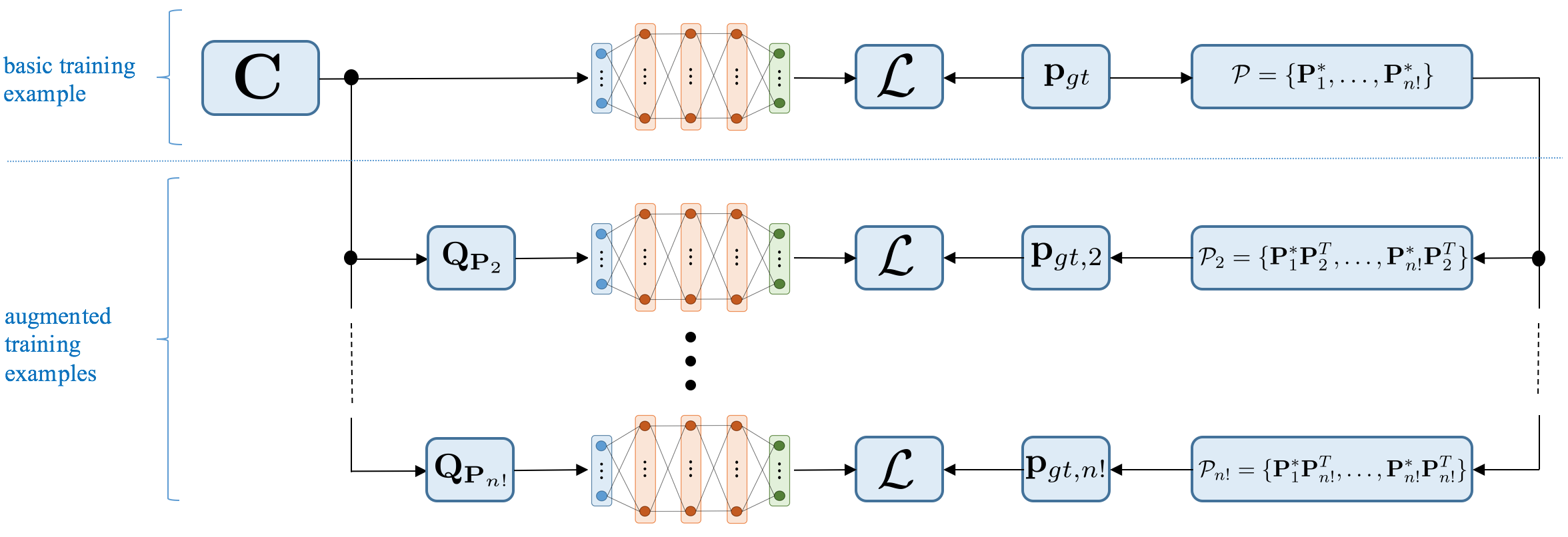}
  \caption{Training data augmentation for permutation-invariant classification performance.}
  \label{fig:augmentation}
\end{figure*}

For each sample coefficient matrix $\mathbf{C}$, the corresponding ground-truth training vector $\mathbf{p}_{gt}$ is then generated by brute-force looping through all possible $n!$ permutations, each time applying the solution strategy outlined in Figure \ref{fig:basicconcept}. The solutions for each permutation are then back-substituted into the original polynomial constraints $f_j$ in order to obtain the mean of the absolute algebraic residuals of each polynomial. The ground-truth vector $\mathbf{p}_{gt}$ is produced by ranking all solutions and thereby distributing the values $\frac{0}{n!-1} \text{ (worst) } \ldots \frac{n!-1}{n!-1} \text{ (best) }$ to each permutation. The training itself minimizes the MSE between $\mathbf{p}$ and $\mathbf{p}_{gt}$, as in $\mathbb{L}=\frac{1}{n!}\|\mathbf{p}-\mathbf{p}_{gt}\|_2^2$. The batch size is set to 128, and we use the Adam solver with parameters set to $lr=0.001$, $\beta_1=0.9$, $\beta_2=0.999$, and $\epsilon = 10^{-8}$.

\subsection{Permutation-invariant classification}

A remaining problem with the classifier of the previous section is that if the optimal permutation for $\{ f_j \}$ is
$\mathbf{P}^*$, and $\pi$ is some permutation represented by matrix $\mathbf{P}$, then the predicted permutation for $\{ \pi(f_j) \}$ need not be $\mathbf{P}^*\mathbf{P}^T$. Note that the latter would have been precisely the case for a permutation invariant classifier. We implicitly enforce the permutation invariance of our classifier via simple training data augmentation.

More precisely, let $\mathbf{Q}_{\mathbf{P}}$ be the permutation induced to the monomials $\mathbf{m}$ by a permutation $\mathbf{P}$. Then there is a permutation $\mathbf{Q}_{\mathbf{P}}^T$ induced from the right to $\mathbf{C}$, so that 
$\mathbf{C} \mathbf{Q}_{\mathbf{P}}^T$ is the coefficient matrix of $\{ \pi(f_j) \}$ with respect to the monomials $\mathbf{Q}_{\mathbf{P}} \mathbf{m}$. Then for each basic training example $\{\mathbf{C},\mathbf{p}_{gt}\}$, we add $n!-1$ further training examples $\left\{\left\{\mathbf{C}\mathbf{Q}^{T}_{\mathbf{P}_{2}},\mathbf{p}_{gt,2}\right\},\ldots,\left\{\mathbf{C}\mathbf{Q}^{T}_{\mathbf{P}_{n!}},\mathbf{p}_{gt,n!}\right\}\right\}$, as outlined in Figure \ref{fig:augmentation}. However, rather than reapplying the above-outlined brute-force search strategy to identify the groundtruth classification results $\{\mathbf{p}_{gt,2},\ldots,\mathbf{p}_{gt,n!}\}$, each one of them is directly and consistently derived from the original groundtruth classification result $\mathbf{p}_{gt}$. This works as follows. We take $\mathbf{p}_{gt}$ and extract a sequence of permutation matrices ordered in decreasing quality, denoted $\mathcal{P}=\{\mathbf{P}^{*}_1,\ldots,\mathbf{P}^{*}_{n!}\}$. For the $i$th augmented training example generated by the permutation matrix $\mathbf{P}_i$, we then extract the consistent ordered sequence $\mathcal{P}_i=\{\mathbf{P}^{*}_1\mathbf{P}_i^{T},\ldots,\mathbf{P}^{*}_{n!}\mathbf{P}_i^{T}\}$. The groundtruth classification result $\mathbf{p}_{gt,i}$ is readily extracted from here.

\section{Results}
\label{sec:results}

We test our method on two generic dense polynomial solvers and one concrete example from geometric vision.

\subsection{Potential improvement on general solvers for dense polynomial systems}

\begin{figure}[t]
  \centering
  \subfigure[]{\label{fig:3301}\includegraphics[width=0.9\linewidth]{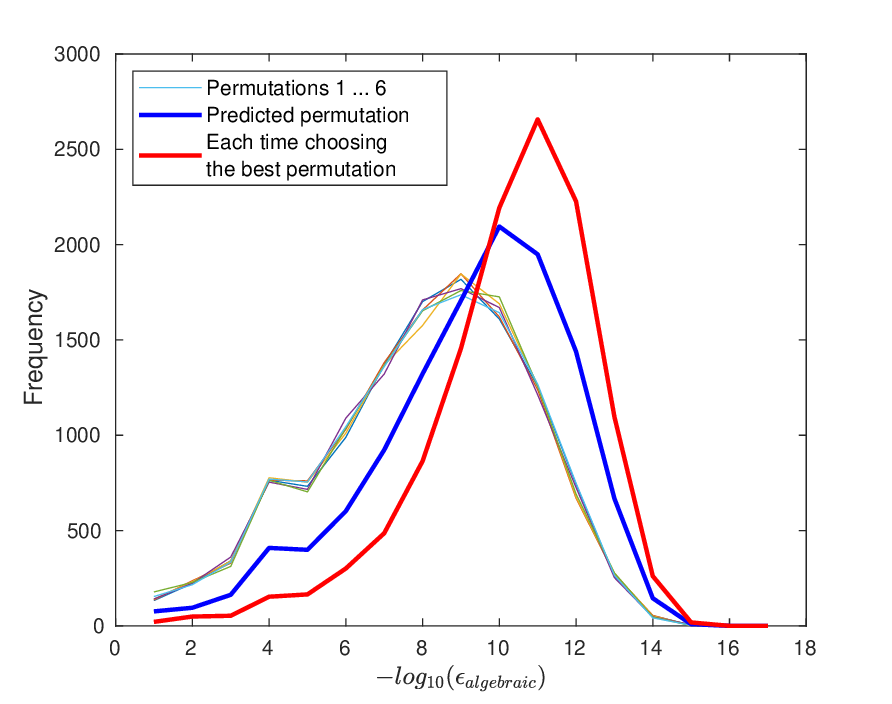}}
  \subfigure[]{\label{fig:4201}\includegraphics[width=0.9\linewidth]{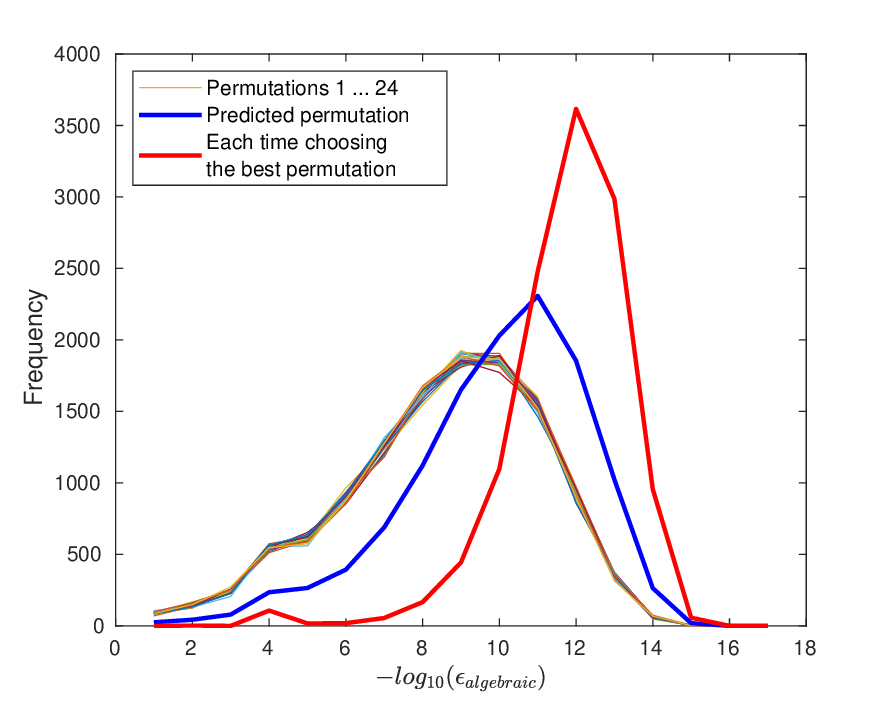}}
  \caption{Error distribution over many random instances of the 3 variable/max-degree 3 (a) and the 4 variable/max-degree 2 (b) problems. Coefficients are randomly chosen from the range [0,1] or [0,10]. Each curve represents the distribution obtained by one of the permutations. The red curve indicates the obtainable result if---for each instance---the best permutation is chosen. The blue curve is the distribution obtained by using the permutation predicted by the classifier.}
  \label{fig:33014201}
\end{figure}

Our first example of a generic dense polynomial problem has three variables and three equations of order three:
\begin{equation*}
    \mathbf{m} = [x_1^3, x_1^2x_2, x_1x_2^2, x_2^3, x_1^2x_3, x_1x_2x_3, x_2^2x_3, x_1x_3^2,
\end{equation*}
\begin{equation*}
    x_2x_3^2, x_3^3, x_1^2, x_1x_2, x_2^2, x_1x_3, x_2x_3, x_3^2, x_1, x_2, x_3, 1]^\mathrm{T} 
\end{equation*}
The improvement in solver stability resulting from the utilization of a permutation classifier is indicated in Figure \ref{fig:3301}. Our second problem has four variables and four equations of maximum degree 2:
\begin{equation*}
    \mathbf{m} = [x_1^2, x_1x_2, x_2^2, x_1x_3, x_2x_3, x_3^2, x_1x_4, x_2x_4,x_3x_4, 
\end{equation*}
\begin{equation*}
    x_4^2, x_1, x_2, x_3, x_4, 1]^\mathrm{T} 
\end{equation*}
The result is indicated in Figure \ref{fig:4201}. As can be observed, adding the classifier enables an improvement of the numerical stability of the solver. The coefficients of $\mathbf{C}$ are chosen randomly from the range [0,1] and [0,10]. Note that there is no fixed pattern in $\mathbf{C}$, the range and the coefficients are chosen fully randomly for each new experiment. Examples with no real solutions are ignored. For examples with multiple solutions, we simply take the average of the sum of absolute residuals of each solution as a ranking or evaluation error. For each problem, we generate $100000 \times n!$ samples, where $76000 \times n!$ are used for training, $12000 \times n!$ for validation, and $12000 \times n!$ for testing. Each case is trained for 200 epochs. In terms of timing, $n!$ evaluations would need 2.44ms and 5.43ms for these two problems respectively. Running the solver once with prior neural network based prediction of a permutation takes only 0.63ms and 0.89ms, respectively. The potential speed-up factor becomes larger as the number of variables is increasing.

\subsection{What works and what not?}

\begin{figure*}[t]
  \centering
  \subfigure[]{\label{fig:2401}\includegraphics[height=0.19\linewidth]{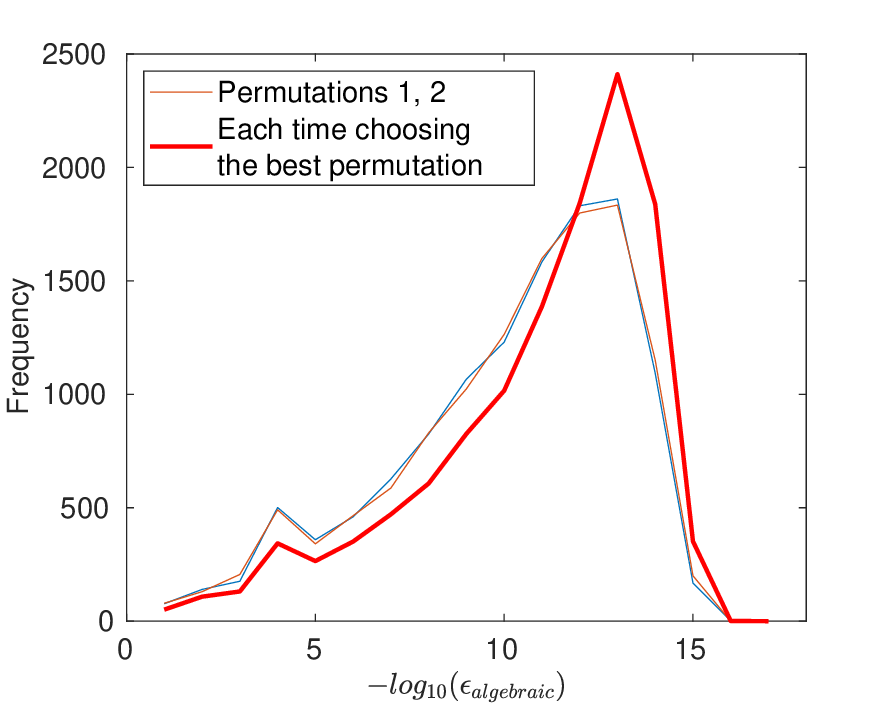}}
  \subfigure[]{\label{fig:33012}\includegraphics[height=0.19\linewidth]{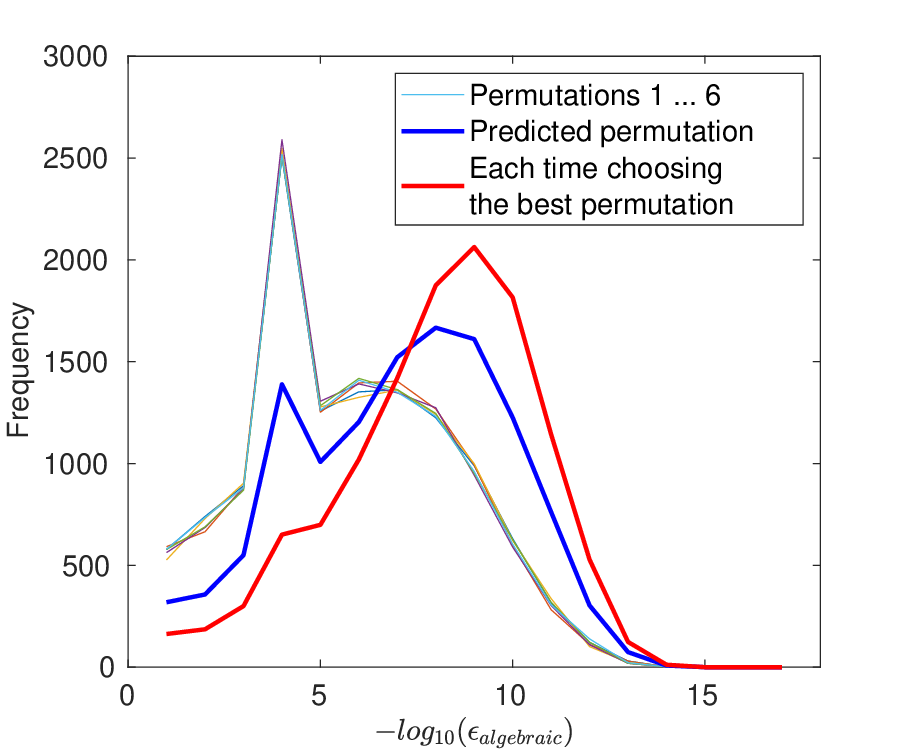}}
  \subfigure[]{\label{fig:42012}\includegraphics[height=0.18\linewidth]{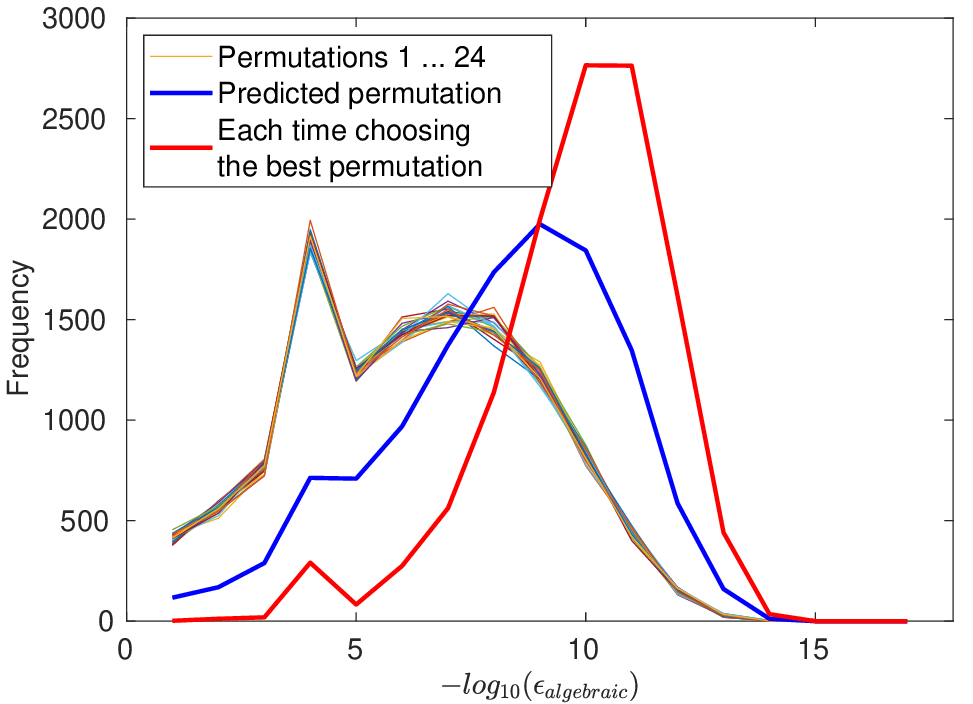}}
  \subfigure[]{\label{fig:420123}\includegraphics[height=0.19\linewidth]{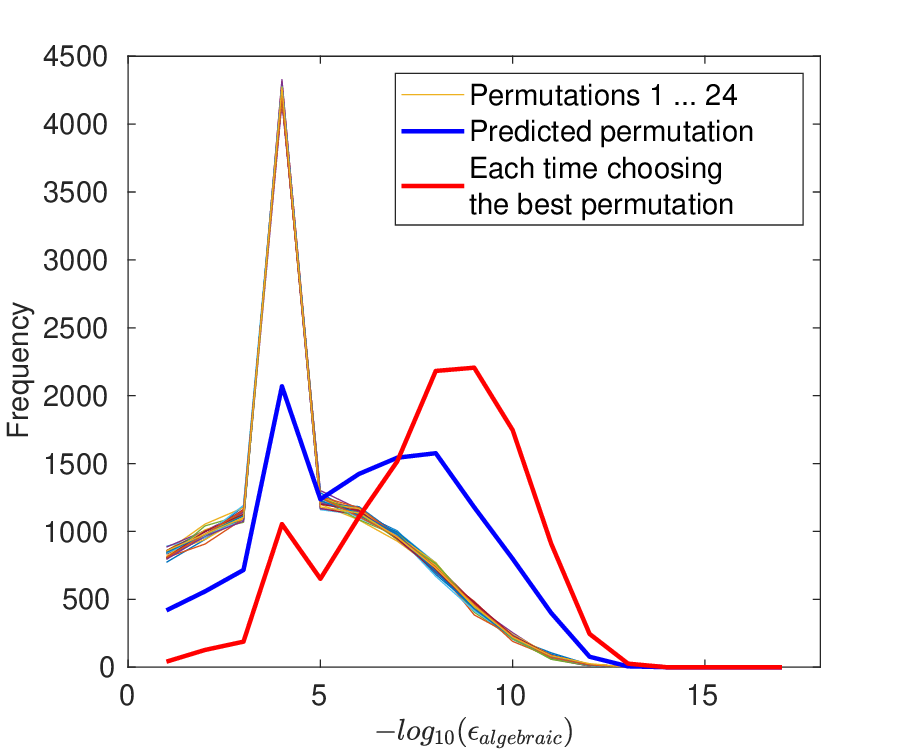}}
  \caption{Error distributions for further cases: order 4 in 2 variables (a), order 3 in 3 variables (b), and order 2 in 4 variables (c) and (d). For (b) and (c), the coefficients are randomly drawn from intervals that are either [0,1], [0,10] or [0,100]. For (d), we also include [0,1000] as a further possible range. Each curve represents the distribution obtained by one of the permutations. Red curves indicate the obtainable result if---for each instance---the best permutation is chosen. Blue curves are the distributions obtained by using the permutation predicted by the classifier.}
  \label{fig:yesandno}
\end{figure*}

The order of the polynomials as well as the number of variables needs to be sufficiently high in order to see substantial differences between the error distributions for each individual permutation or each time choosing the best one. For example, Figure \ref{fig:2401} shows the distributions for a quadratic problem in 4 unknowns, indicating that the application of a classifier would not lead to any substantial benefits. What furthermore matters is the presence of sufficient structure in the problem. Our first test consisted of simply choosing the same order of magnitude for all random coefficients, which did not lead to any successful outcomes. It can be concluded that numerical stability only becomes predictable if the coefficients cover a sufficiently large spectrum of possible values. To illustrate this further, we have modified both cases outlined in the previous section to include coefficients from three different ranges, i.e. [0,1], [0,10], and [0,100]. For the four-variable problem, we furthermore ran one further experiment where we include yet another range into the random coefficient generation, i.e. [0,1000]. The results are indicated in Figured \ref{fig:33012}, \ref{fig:42012}, and \ref{fig:420123}, respectively. As can be observed, the gap between the distribution over individual permutations and the best permutation is increasing along with the possible order of magnitude for the coefficients, and the permutation chosen by the classifier leads to a more pronounced improvement.

\subsection{Improvement of camera resectioning algorithm}

Our final experiment consists of an application to a state-of-the-art camera resectioning algorithm, the UPnP algorithm by Kneip et al. UPnP\cite{kneip2014upnp}. The goal of the algorithm consists of using an arbitrary number of correspondences between 2D image point measurements and 3D world point coordinates to calculate the six degree-of-freedom absolute pose of the camera. The algorithm makes the assumption of known intrinsic camera parameters, and we furthermore apply it in the central, single camera case. We choose small but non-minimal numbers of correspondences in each experiment, and do not add any noise in order to properly evaluate numerical accuracy.

The UPnP algorithm first eliminates the translation parameters from the estimation, and aims at solving the first-order optimality conditions given by
\begin{equation}
\mathrm{E} =\tilde{\mathbf{s}}^\mathrm{T}\mathbf{C}\tilde{\mathbf{s}},
\label{equ:sMs}
\end{equation}
where, $\tilde{\mathbf{s}} = [ \mathbf{s}^\mathrm{T}\ 1]^\mathrm{T}$, and
\begin{equation*}
    \mathbf{s} = [q_0^2, q_1^2, q_2^2, q_3^2, q_0q_1, q_0q_2, q_0q_3, q_1q_2, q_1q_3, q_2q_3]^\mathrm{T}
\end{equation*}
represents a vector of all second-order monomials of the quaternion parameters $\mathbf{q}=[q_0, q_1, q_2, q_3]^T$. Completed by a unit-norm constraint on $\mathbf{q}$, the final polynomial problem is permutation invariant and of order 3 in 4 variables.

\begin{figure}[b]
\begin{center}
 \includegraphics[width=0.9\linewidth]{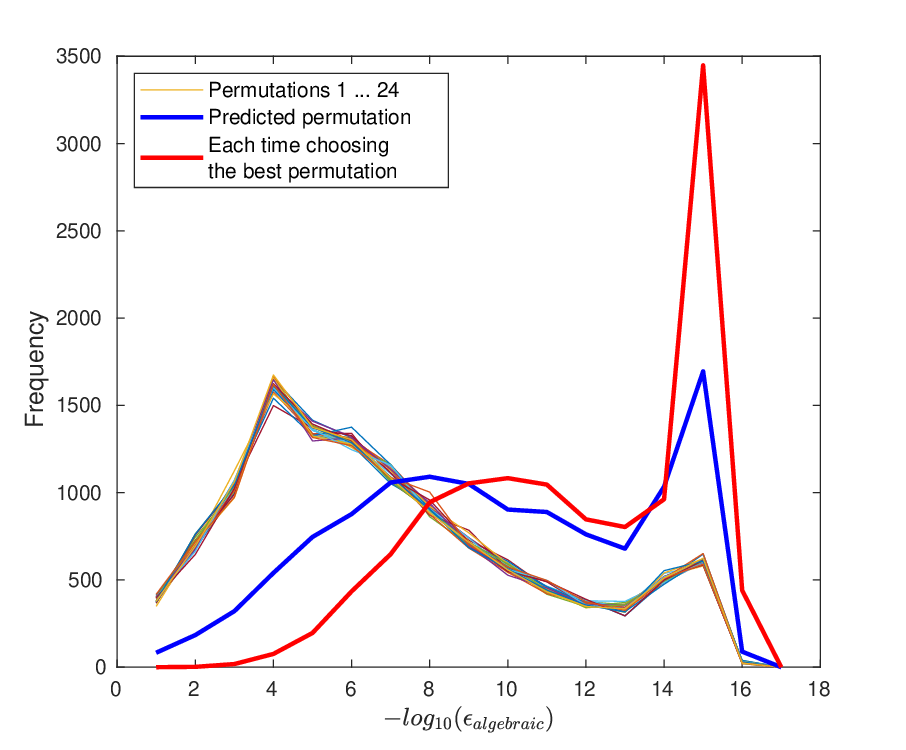}
\end{center}
  \caption{Error distribution over many random instances of the UPnP problem \cite{kneip2014upnp}. Each curve represents the distribution obtained by one of the permutations. The red curve furthermore indicates the obtainable result if---for each instance---the best permutation is chosen. The blue curve is the distribution of the predicted permutation.}
\label{fig:final}
\end{figure}

The input to our system is a 55-dimensional vector. We train the classifier by generating a total of $100000 \times 4!$ samples, and use $76000 \times 4!$ samples for training, $12000 \times 4!$ for validation, and $12000 \times 4!$ for testing. We run a total of 200 epochs. Working on a geometric solver lets us furthermore replace the algebraic residual errors by $\left\| \mathbf{T}-\mathbf{T}_{\text{gt}} \right\|_\mathbf{Frob}$, where $\mathbf{T}$ and $\mathbf{T}_{gt}$ represent the calculated and the groundtruth pose. Our result is indicated in Figure \ref{fig:final} and demonstrates how the addition of the classifier contributes to a significant increase in the numerical stability and accuracy of the solver. We deem this result as quite important. First, it shows that geometric problems potentially contain the necessary structure in the coefficients. Second, many solvers from geometric vision parametrize the problem as a function of the rotation, and thus appear in permutation invariant form and may potentially benefit from the addition of a similar classifier.

\section{Discussion}
\label{sec:conclusion}

Our paper demonstrates two main ideas. First, we show that for certain types of problems there may be multiple ways to execute an elimination template, and notably with similar computational efficiency but different numerical stability. In particular, we define permutation invariant polynomial systems as a large class of problems for which such a choice is easily achieved by simple permutations. Second, we prove that the original coefficients can be used to predict a good permutation, thus enabling a cost-effective improvement of the numerical stability of Gr\"obner basis solvers. The significance of our contribution is increased by the fact that the class of permutation invariant polynomial problems include many solvers from the field of geometric vision, most notably all solvers that apply a polynomial parametrization of rotation matrices, such as Cayley or quaternion parameters. It is also easy to recognize that the method can be transparently extended to problems that contain permutation-variant polynomials, but remain permutation invariant from the point of the view of the entire system. A further applicable case is when only a subset of the variables or monomials appears in permutation invariant form. Our future work therefore consists of extending the idea to more general cases including not only such partially permutation invariant problems, but also cases for which simply different, similarly efficient elimination templates exist.

\section*{Acknowledgements}

The authors would like to acknowledge the generous startup funds 2017F0203-000-15 and 2017F0203-000-16 provided by ShanghaiTech University and the Chinese Academy of Sciences.

{\small
\bibliographystyle{ieee}
\bibliography{egbib}

\begin{thebibliography}{10}\itemsep=-1pt

\bibitem{ask12}
E.~Ask, K.~Yubin, and K.~Astrom.
\newblock Exploiting p-fold symmetries for faster polynomial equation solving.
\newblock In {\em Proceedings of the International Conference on Pattern
  Recognition ({ICPR})}, 2012.

\bibitem{buchberger85}
B.~Buchberger.
\newblock {\em Multidimensional Systems Theory - Progress, Directions and Open
  Problems in Multidimensional Systems}.
\newblock Reidel Publishing Company, Dodrecht - Boston - Lancaster, 1985.

\bibitem{bujnak2012making}
M.~Bujnak, Z.~Kukelova, and T.~Pajdla.
\newblock Making minimal solvers fast.
\newblock In {\em Proceedings of the IEEE Conference on Computer Vision and
  Pattern Recognition (CVPR)}, 2012.

\bibitem{byrod09}
M.~Byr\"{o}d, K.~Josephson, and K.~Astrom.
\newblock Fast and stable polynomial equation solving and its application to
  computer vision.
\newblock {\em International Journal of Computer Vision}, 84(3):237--256, 2009.

\bibitem{cox2006using}
D.~Cox, J.~Little, and D.~O'Shea.
\newblock {\em Using algebraic geometry}, volume 185.
\newblock Springer Science \& Business Media, 2006.

\bibitem{cox1994ideals}
D.~Cox, J.~Little, D.~O'Shea, and M.~Sweedler.
\newblock Ideals, varieties, and algorithms.
\newblock {\em American Mathematical Monthly}, 101(6):582--586, 1994.

\bibitem{faugere93}
J.~Faug\`ere, P.~Gianni, D.~Lazard, and T.~Mora.
\newblock {Efficient Computation of Zero-dimensional Gr\"obner Bases by Change
  of Ordering}.
\newblock {\em Journal of Symbolic Computation}, 16(4):329--344, 1993.

\bibitem{kneip2014upnp}
L.~Kneip, H.~Li, and Y.~Seo.
\newblock Upnp: An optimal o (n) solution to the absolute pose problem with
  universal applicability.
\newblock In {\em Proceedings of the European Conference on Computer Vision
  (ECCV)}, pages 127--142. Springer, 2014.

\bibitem{kneip12}
L.~Kneip, R.~Siegwart, and M.~Pollefeys.
\newblock {F}inding the {E}xact {R}otation {B}etween {T}wo {I}mages
  {I}ndependently of the {T}ranslation.
\newblock In {\em Proceedings of the European Conference on Computer Vision
  (ECCV)}, Firenze, Italy, 2012.

\bibitem{kukelova14}
Z.~Kukelova, M.~Bujnak, J.~Heller, and T.~Pajdla.
\newblock Singly-bordered block-diagonal form for minimal problem solvers.
\newblock In {\em Proceedings of the Asian Conference on Computer Vision
  (ACCV)}, 2014.

\bibitem{kukelova2008automatic}
Z.~Kukelova, M.~Bujnak, and T.~Pajdla.
\newblock Automatic generator of minimal problem solvers.
\newblock In {\em Proceedings of the European Conference on Computer Vision
  (ECCV)}, pages 302--315, 2008.

\bibitem{kukelova13}
Z.~Kukelova, M.~Bujnak, and T.~Pajdla.
\newblock Real-time solution to the absolute pose problem with unknown radial
  distortion and focal length.
\newblock In {\em Proceedings of the IEEE International Conference on Computer
  Vision (ICCV)}, 2013.

\bibitem{larsson2017efficient}
V.~Larsson, K.~Astrom, and M.~Oskarsson.
\newblock Efficient solvers for minimal problems by syzygy-based reduction.
\newblock In {\em Proceedings of the IEEE Conference on Computer Vision and
  Pattern Recognition (CVPR)}, pages 820--829, 2017.

\bibitem{larsson17iccv2}
V.~Larsson, K.~Astrom, and M.~Oskarsson.
\newblock Polynomial solvers for saturated ideals.
\newblock In {\em Proceedings of the IEEE International Conference on Computer
  Vision (ICCV)}, 2017.

\bibitem{larsson2018beyond}
V.~Larsson, M.~Oskarsson, K.~Astrom, A.~Wallis, Z.~Kukelova, and T.~Pajdla.
\newblock Beyond gr\"obner bases: Basis selection for minimal solvers.
\newblock In {\em Proceedings of the IEEE Conference on Computer Vision and
  Pattern Recognition (CVPR)}, pages 3945--3954, 2018.

\bibitem{lee14}
G.~Lee, M.~Pollefeys, and F.~Fraundorfer.
\newblock Relative pose estimation for a multi-camera system with known
  vertical.
\newblock In {\em Proceedings of the IEEE Conference on Computer Vision and
  Pattern Recognition (CVPR)}, 2014.

\bibitem{mora88}
T.~Mora and L.~Robbiano.
\newblock The gr\"obner fan of an ideal.
\newblock {\em Journal of Symbolic Computation}, 6(2--3):183--208, 1988.

\bibitem{peng19}
L.~Peng, X.~Song, M.~C. Tsakiris, H.~Choi, L.~Kneip, and Y.~Shi.
\newblock Algebraically-initialized expectation maximization for header-free
  communication.
\newblock In {\em Proceedings of the {IEEE} International Conference on
  Accoustics, Speech and Signal Processing ({ICASSP})}, 2019.

\bibitem{cruz18}
R.~Santa~Cruz, B.~Fernando, A.~Cherian, and S.~Gould.
\newblock Visual permuation learning, 2018.

\bibitem{stewenius2006recent}
H.~Stew\'enius, C.~Engels, and D.~Nist{\'e}r.
\newblock Recent developments on direct relative orientation.
\newblock {\em ISPRS Journal of Photogrammetry and Remote Sensing},
  60(4):284--294, 2006.

\bibitem{stewenius05}
H.~Stew\'enius and D.~Nist\'er.
\newblock {S}olutions to {M}inimal {G}eneralized {R}elative {P}ose {P}roblems.
\newblock In {\em Workshop on Omnidirectional Vision (ICCV)}, Beijing, China,
  2005.

\bibitem{tsakiris18}
M.~C. Tsakiris, L.~Peng, A.~Conca, L.~Kneip, Y.~Shi, and H.~Choi.
\newblock An algebraic-geometric approach to shuffled linear regression.
\newblock {\em {ArXiv e-prints}}, 2018.

\bibitem{zhao18}
J.~Zhao, L.~Kneip, Y.~He, and J.~Ma.
\newblock Minimal case relative pose computation using ray-point-ray features.
\newblock {\em Transactions of Pattern Analysis and Machine Intelligence},
  2018.

\bibitem{zheng13}
Y.~Zheng, Y.~Kuang, S.~Sugimoto, K.~Astrom, and M.~Okutomi.
\newblock Revisiting the {P}n{P} problem: A fast, general and optimal solution.
\newblock In {\em Proceedings of the IEEE International Conference on Computer
  Vision (ICCV)}, 2013.

\end{thebibliography}
}

\end{document}